\documentclass[a4paper]{article}

\usepackage[numbers,sort&compress]{natbib}
\usepackage{booktabs,url}
\usepackage[preprint]{INTERSPEECH2018}
\usepackage{bm}

\title{Semi-tied Units for Efficient Gating in LSTM and Highway Networks}
\name{{C. Zhang \& P. C. Woodland}\thanks{Thanks to Mark Gales \& the MGB3 team for the MGB3 setup used.}}
\address{Cambridge University Engineering Dept., Trumpington St., Cambridge, CB2 1PZ U.K.}
\email{\small \tt \{cz277,pcw\}@eng.cam.ac.uk}
  
\begin{document}

\maketitle
\begin{abstract}
Gating is a key technique used for integrating information from multiple sources by long short-term memory (LSTM) models and has recently also been applied to other models such as the highway network. Although gating is powerful, it is rather expensive in terms of both computation and storage as each gating unit uses a separate full weight matrix. This issue can be severe since several gates can be used together in e.g. an LSTM cell. This paper proposes a semi-tied unit (STU) approach to solve this efficiency issue, which uses one shared weight matrix to replace those in all the units in the same layer. The approach is termed ``semi-tied'' since extra parameters are used to separately scale each of the shared output values.  These extra scaling factors are associated with the network activation functions and result in the use of parameterised sigmoid, hyperbolic tangent, and rectified linear unit functions. Speech recognition experiments using British English multi-genre broadcast data showed that using STUs can reduce the calculation and storage cost  by a factor of  three for highway networks and four for LSTMs, while giving similar word error rates to the original models.

\end{abstract}

\toappear{To appear in Proc. INTERSPEECH 2018, September 2-6, 2018, Hyderabad, India}
\copyrightnotice{\copyright 2018 ISCA}

\section{Introduction}
Gating units have become a key component in many types of artificial neural network (ANN) models.  These units yield soft 0-1 valued outputs that are used to scale signals from other parts of the network.
In recurrent neural networks (RNNs) a key issue  is solving the  vanishing gradient problem \cite{Bengio:1994ab} in training
which causes   standard RNNs to find it difficult to learn long-term information. Gated RNNs such as the long short-term memory (LSTM) model   \cite{Hochreiter:1997ab} define explicit memory cells where updates to the
memory cell values are controlled by two gating units and updating the hidden state value uses a further gate. 
The gated recurrent unit (GRU) is an RNN that uses two gates \cite{Chung:2014ab}. 

Recently, highway connections have been proposed to enable a feed-forward or a recurrent layer to have an extra nonlinearity by combining its input and output values via gating units \cite{Srivastava:2015ab,Srivastava:2015cd,Srivastava:2016ab}. The highway idea has also been applied to connect the memory cells of neighbouring LSTM layers \cite{Yao:2015ab}. Furthermore, gating is also useful for convolutional layers \cite{Shi:2015ab,Bradbury:2017ab}. 
 A quasi-RNN uses gates to integrate different time step outputs from a layer shared across time, which can be viewed as temporal convolutional model as the shared layer serves as a time-invariant filter \cite{Bradbury:2017ab}. 
All models discussed above have been applied to acoustic modelling for speech recognition \cite{Graves:2009ab,Graves:2013ab,Sak:2014ab,Sak:2015ab,Zhang:2016ab,Lu:2016cd,Pundak:2017ab,Lu:2016ab,Zhang:2017cd,Tao:2017ab}.

Normally a gating unit is defined as a sublayer that outputs a ``gating vector'' of soft 0-1 values by operating on e.g. current input values or those from previous layers, with full weight matrices. 
This gating vector is often
applied to a ``candidate vector'' that would, for instance in the case of an LSTM, be used to update the memory cell values. Since the calculation of the gating vectors often has a similar functional form to that used to find the candidate vectors,
the overall number of parameters and computational complexity of gated models is high \cite{Hochreiter:1997ab,Shi:2015ab,Srivastava:2015ab}. This can be severe when models use several different gating units.

In this paper we propose an alternative type of unit for gating termed a semi-tied unit (STU), which aims at implementing a similar function to the traditional gating unit in a more efficient way. As its name suggests, the key idea in STU is to share parameters to save computation while also adding some untied parameters so that gating units can learn distinct functions. 
This paper studies the most commonly used gated models, LSTMs and highway networks, which have each of their units implemented based on full weight matrices.
In order to reduce the number of matrix multiplications, the STUs share the weights and biases among all  the gating and candidate units in the same layer. Meanwhile, additional untied parameter vectors are introduced as component-wise adaptive scaling factors through 
parameterised activation functions \cite{Zhang:2015cd,Zhang:2017ab}, which allows the STUs to generate distinct gating and candidate vectors. 
Experimental results found using STUs in both LSTM and highway network resulted in similar WERs to those based on traditional gating units, while being significantly more efficient.

The rest of the paper is organised as follows. Section 2 reviews the gating mechanism along with LSTMs and highway networks. STUs based on parameterised activation functions are described in Section 3. The experimental setup and results are given in Sections 4 and 5, which is followed by conclusions.

\section{Gating Mechanism}
Analogous to an array of logic gate in electronics, an ANN gating unit converts its vector input into a 0-1 valued gating vector. For an LSTM layer at time $t$, the input, forget, and output gating vectors $\mathbf{i}_t$, $\mathbf{f}_t$, and $\mathbf{o}_t$ are computed by
\begin{align}
	\label{eq:1}
	\mathbf{i}_t&=\sigma(\mathbf{W}_i\mathbf{x}_t+\mathbf{U}_i\mathbf{h}_{t-1}+\mathbf{V}_i\circ\mathbf{c}_{t-1}+\mathbf{b}_i)\\
	\label{eq:2}
	\mathbf{f}_t&=\sigma(\mathbf{W}_f\mathbf{x}_t+\mathbf{U}_f\mathbf{h}_{t-1}+\mathbf{V}_f\circ\mathbf{c}_{t-1}+\mathbf{b}_f)\\
	\label{eq:3}
	\mathbf{o}_t&=\sigma(\mathbf{W}_o\mathbf{x}_t+\mathbf{U}_o\mathbf{h}_{t-1}+\mathbf{V}_o\circ\mathbf{c}_{t}+\mathbf{b}_o),
\end{align}
where $\mathbf{x}_t$ and $\mathbf{h}_{t}$ are the input and hidden state values; $\mathbf{W}$ and $\mathbf{U}$ are weight matrices; $\mathbf{b}$ and $\mathbf{V}$ are the  bias vector and diagonal ``peephole'' matrix; $\circ$ is component-wise multiplication; $[\sigma(\mathbf{a_t})]_j=1/(1+e^{-{a}_{tj}})$ is the $j^\text{th}$ component of the  \textit{vector sigmoid function} with input activation vector $\mathbf{a}_t$, and ${a}_{tj}$ is the $j^\text{th}$ component of $\mathbf{a}_t$.
Given the $j^\text{th}$ component of the \textit{vector hyperbolic tangent function} as $[\tanh(\mathbf{a}_{t})]_j =(e^{{a}_{tj}}-e^{-{a}_{tj}})/(e^{{a}_{tj}}+e^{-{a}_{tj}})$, the gating vectors are used to generate $\mathbf{h}_t$ based on the previous memory cell value $\mathbf{c}_{t-1}$ and 
the current candidate vector $\tilde{\mathbf{c}}_{t}$ by 
\begin{align}
	\label{eq:4}
	\tilde{\mathbf{c}}_t&=\text{tanh}(\mathbf{W}_c\mathbf{x}_t+\mathbf{U}_c\mathbf{h}_{t-1}+\mathbf{b}_c)\\
	\label{eq:5}
	\mathbf{c}_t&=\mathbf{f}_t\circ\mathbf{c}_{t-1}+\mathbf{i}_t\circ\tilde{\mathbf{c}}_t\\
	\label{eq:6}
	\mathbf{h}_t&=\mathbf{o}_t\circ\text{tanh}(\mathbf{c}_t),
\end{align}
which manipulates the information flow to simulate the human long short-term memory mechanism, and helps solve the gradient vanishing problem \cite{Hochreiter:1997ab}. 

The gating idea can also be used to attenuate the information loss in feedforward layers to allow the training of very deep models \cite{Srivastava:2015cd}.
 A \textit{highway network} refers to a feedforward model with a stack of highway layers \cite{Srivastava:2015ab}, with each of them defined as
\begin{align}
	\label{eq:7}
	\mathbf{m}_t&=\sigma(\mathbf{W}_m\mathbf{x}_t+\mathbf{b}_m)\\
	\label{eq:8}
	\mathbf{r}_t&=\sigma(\mathbf{W}_r\mathbf{x}_t+\mathbf{b}_r)\\
	\label{eq:9}
	\tilde{\mathbf{y}}_t&=f(\mathbf{W}_y\mathbf{x}_t+\mathbf{b}_y)\\
	\label{eq:10}
	\mathbf{y}_t&=\mathbf{m}_t\circ\tilde{\mathbf{y}}_t+\mathbf{r}_t\circ\mathbf{x}_t,
\end{align}
where $\mathbf{y}_t$ is the output of the layer, $\tilde{\mathbf{y}}_t$ is the candidate vector,  $\mathbf{m}_t$ and $\mathbf{r}_t$ are the gating vectors of the transform and carry gates, and $f(\cdot)$ is an activation function. For recurrent models, the feedforward highway layers can be located in-between the recurrent layers, which results in the recurrent highway network \cite{Srivastava:2016ab}. 
Furthermore, $\mathbf{r}_t$ can be replaced by $\mathbf{1}-\mathbf{m}_t$ to save one gating unit in each layer, and Eqn.~\eqref{eq:10} becomes
\begin{align}
	\label{eq:11}
	\mathbf{y}_t&=\mathbf{m}_t\circ\tilde{\mathbf{y}}_t+(\mathbf{1}-\mathbf{m}_t)\circ\mathbf{x}_t.
\end{align}
This idea has also been applied to GRUs and quasi-RNNs by modifying Eqn.~\eqref{eq:6} in the same way \cite{Chung:2014ab,Bradbury:2017ab}. 
However, since Eqn.~\eqref{eq:6} is found to work better for highway networks \cite{Lu:2016ab}, it is used throughout this paper.

\section{Semi-tied Units}
From Eqns.~\eqref{eq:1} -- \eqref{eq:4} and Eqns.~\eqref{eq:7} -- \eqref{eq:9}, only a quarter or one third of the parameters (and calculations) are used to generate the candidate vectors in an LSTM and highway layers respectively, while the rest are associated with gating. The efficiency could be improved if there exists a shared ``virtual unit'' which distinguishes the gating and candidate units  by cheaper operations than matrix multiplications. This is reasonable since the units have the same input and functional form. Based on this assumption, the STU is proposed that represents the ``virtual unit'' by parameters that are tied across all gating and candidate units, and modelling the difference between the ``virtual unit'' and every other unit by some extra untied parameters.

\subsection{Parameterised Activation Function for STUs}
In LSTMs and highway networks, since weight matrix multiplications take the most computation and storage cost, they are tied to form the ``virtual unit'', and the bias vectors are also tied.
 The type of the untied parameters is another important choice in an STU as they model the differences between the units. This paper uses additional linear factors to scale the output values from the ``virtual unit'' for this purpose, which is very efficient as it involves only component-wise operations. It is natural to associate such scaling factors with the activation functions that leads to the use of the parameterised activation functions proposed in \cite{Zhang:2015cd}. The parameterised sigmoid function with additional parameter vectors $\bm{\eta}$ and $\bm{\gamma}$ is denoted as $\sigma_{\bm{\eta},\bm{\gamma}}(\cdot)$ and defined by
\begin{align*}
	\sigma_{\bm{\eta},\bm{\gamma}}(\mathbf{a}_t)=\bm{\eta} \circ \sigma(\bm{\gamma} \circ \mathbf{a}_t),
\end{align*}
where $\bm{\eta}$ and $\bm{\gamma}$ associates\ an independent parameter for every output node to scale its output and input values. Note that the scaling by $\bm{\eta}$ can mean that  the range of the  gating vector values is no longer constrained by 0-1, which can be seen as a generalisation of the original gating mechanism. 
In order to use STUs for LSTMs and rectified linear unit (ReLU) highway networks, the $\text{tanh}$ and ReLU functions are also parameterised as 
\begin{align*}
	\text{tanh}_{\bm{\eta},\bm{\gamma}}(\mathbf{a}_t)&=\bm{\eta}\circ\text{tanh}(\bm{\gamma}\circ\mathbf{a}_t)\\
	\text{ReLU}_{\bm{\eta}}(\mathbf{a}_t)&=\bm{\eta}\circ \text{ReLU}(\mathbf{a}_t),
\end{align*}
where $[\text{ReLU}(\mathbf{a}_{t})]_j=\max(a_{tj},0)$.
Here $\bm{\eta}$ and $\bm{\gamma}$ still refer to the output and input value scaling vectors for $\text{tanh}_{\bm{\eta},\bm{\gamma}}(\cdot)$ and $\text{ReLU}_{\bm{\eta}}(\cdot)$. 
Other types of parameterised activation functions have also been investigated for both conventional modelling \cite{Goh:2003ab,Siniscalchi:2010ab,He:2015ab,Tuske:2015ab} and speaker adaptation \cite{Siniscalchi:2013ab,Zhao:2015ab,Siniscalchi:2016ab,Zhang:2016cd}. 

\subsection{STUs for LSTMs and Highway Networks}
\subsubsection{STU based LSTMs ($\text{LSTM}^{\text{STU}}$)}
\label{sec:LSTMSTU}
As discussed before, the weights and biases are tied across all gating and candidate units when using STUs. The shared part, or the ``virtual unit'', produces the common values $\mathbf{e}_t$ by
\begin{align*}
	\mathbf{e}_t&=\mathbf{W}\mathbf{x}_t+\mathbf{U}\mathbf{h}_{t-1}+\mathbf{b}.
\end{align*}
Let $i$, $f$, $o$, and $c$ be the subscripts of the activation function parameters for the input gate, forget gate, output gate, and the candidate unit, Eqns.~\eqref{eq:1} -- \eqref{eq:4} can be re-written as
\begin{align*}
	\mathbf{i}_t&=\sigma_{\bm{\eta}_i,\bm{\gamma}_i}(\mathbf{e}_t+\mathbf{V}\circ\mathbf{c}_{t-1})\\
	\mathbf{f}_t&=\sigma_{\bm{\eta}_f,\bm{\gamma}_f}(\mathbf{e}_t+\mathbf{V}\circ\mathbf{c}_{t-1})\\
	\mathbf{o}_t&=\sigma_{\bm{\eta}_o,\bm{\gamma}_o}(\mathbf{e}_t+\mathbf{V}\circ\mathbf{c}_{t})\\
	\tilde{\mathbf{c}}_t&=\text{tanh}_{\bm{\eta}_c,\bm{\gamma}_c}(\mathbf{e}_t).
\end{align*}
Hence the weight matrices are tied to $\mathbf{W}$ and $\mathbf{U}$ respectively; the bias vectors and diagonal ``peephole'' matrices are tied to $\mathbf{b}$ and $\mathbf{V}$. 
Let $X$ and $H$ be the sizes of $\mathbf{x}_t$ and $\mathbf{h}_t$, STUs reduced the computation and storage complexities from $\mathcal{O}(4XH+4H^2)$ to $\mathcal{O}(XH+H^2)$.
Compared to the projected LSTM (LSTMP), whose recurrent matrices are factorised by a $H\times P$ projection matrix $\mathbf{P}$ to $\mathbf{U}_i\mathbf{P}$, $\mathbf{U}_f\mathbf{P}$, $\mathbf{U}_o\mathbf{P}$, and $\mathbf{U}_c\mathbf{P}$ \cite{Sak:2014ab}, $\text{LSTM}^{\text{STU}}$ is even more efficient than LSTMP with $P=H/4$. The LSTMP also falls into the STU framework, by defining $\mathbf{e}_t=\mathbf{P}\mathbf{h}_{t-1}$.

\subsubsection{STU based Highway Network ($\text{Highway}^{\text{STU}}$)}
Similar to the $\text{LSTM}^{\text{STU}}$ case, the ``virtual unit'' output value $\mathbf{e}_t$ in $\text{Highway}^{\text{STU}}$ is also shared among all gating units and the candidate unit. By tying both weights and biases, we have
\begin{align*}
	\mathbf{e}_t=\mathbf{W}\mathbf{x}_t+\mathbf{b},
\end{align*}
which is equal to the shared input activation values. 

In a sigmoid highway network, i.e. $f(\cdot)=\sigma(\cdot)$, associating $\bm{\eta}_m$ and $\bm{\gamma}_m$ with the transform gate, $\bm{\eta}_r$ and $\bm{\gamma}_r$ with the carry gate, and $\bm{\eta}_y$ and $\bm{\gamma}_y$ with the candidate unit, Eqns.~\eqref{eq:7} -- \eqref{eq:9} are modified as
\begin{align*}
	\mathbf{m}_t&=\sigma_{\bm{\eta}_m,\bm{\gamma}_m}(\mathbf{e}_t)\\
	\mathbf{r}_t&=\sigma_{\bm{\eta}_r,\bm{\gamma}_r}(\mathbf{e}_t)\\
	\tilde{\mathbf{y}}_t&=\sigma_{\bm{\eta}_y,\bm{\gamma}_y}(\mathbf{e}_t).
\end{align*}
This ties all weight matrices and bias vectors together, and reduces the calculation and storage complexities from $\mathcal{O}(3XH)$ to $\mathcal{O}(XH)$. If $f(\cdot)$ is ReLU, Eqn.~\eqref{eq:9} is then replaced by
\begin{align*}
	\tilde{\mathbf{y}}_t&=\text{ReLU}_{\bm{\eta}_y}(\mathbf{e}_t).
\end{align*}	
where $\bm{\eta}_y$ is the ReLU output scaling factor vector. Note in both $\text{LSTM}^{\text{STU}}$ and $\text{Highway}^{\text{STU}}$, almost all parameters and calculations are used in candidate vector generation.

\subsection{Training STUs}

\subsubsection{Training Activation Function Parameters}
To train STUs by \textit{error back propagation}, the derivatives of the parameterised activation functions w.r.t. to the function parameters and input activation values are required \cite{Zhang:2015cd}. 
Let $a_{tj}$, $\eta_j$, and $\gamma_j$ be the $j^{\text{th}}$ components of $\mathbf{a}_t$, $\bm{\eta}$, and $\bm{\gamma}$, 
then 
\begin{align*}
	{\partial\sigma_{\bm{\eta},\bm{\gamma}}(\mathbf{a}_t)}/{\partial a_{tj}}&=\eta_j\gamma_j\, e^{-\gamma_ja_{tj}}/(1+e^{-\gamma_ja_{tj}})^2\\
	{\partial \sigma_{\bm{\eta},\bm{\gamma}}(\mathbf{a}_t)}/{\partial \eta_j}&=1/(1+e^{-\gamma_ja_{tj}})\\
	{\partial \sigma_{\bm{\eta},\bm{\gamma}}(\mathbf{a}_t)}/{\partial \gamma_j}&=\eta_ja_{tj}e^{-\gamma_ja_{tj}}/(1+e^{-\gamma_ja_{tj}})^2,
\end{align*}
and
\begin{align*}
	{\partial\,\text{tanh}_{\bm{\eta},\bm{\gamma}}(\mathbf{a}_t)}/{\partial a_{tj}}&=4\eta_j\gamma_j/(e^{\gamma_ja_{tj}}+e^{-\gamma_ja_{tj}})^2\\
	{\partial\,\text{tanh}_{\bm{\eta},\bm{\gamma}}(\mathbf{a}_t)}/{\partial\eta_j}&=(e^{\gamma_ja_{tj}}-e^{-\gamma_ja_{tj}})/(e^{\gamma_ja_{tj}}+e^{-\gamma_ja_{tj}})\\
	{\partial\,\text{tanh}_{\bm{\eta},\bm{\gamma}}(\mathbf{a}_t)}/{\partial\gamma_j}&=4\eta_ja_{tj}/(e^{\gamma_ja_{tj}}+e^{-\gamma_ja_{tj}})^2.
\end{align*}
Similarly, for $\text{ReLU}_{\bm{\eta}}(\mathbf{a}_t)$, we have
\begin{align*}
	{\partial\,\text{ReLU}_{\bm{\eta}}(\mathbf{a}_t)}/{\partial a_{tj}}&=\left\{\begin{array}{ll}
      0  & \text{if}{~~} a_{tj}<0\\
   	  \eta_j & \text{if}{~~} a_{tj}\geqslant0
   	  \end{array}\right.\\
	{\partial\,\text{ReLU}_{\bm{\eta}}(\mathbf{a}_t)}/{\partial\eta_j}&=\max(a_{tj},0).
\end{align*}	

\subsubsection{Normalising the Gradients of the Tied Parameters}
When training the shared parameters in the ``virtual unit'' of an STU, e.g. $\mathbf{W}$, there are
\begin{align*}
	\dfrac{\partial\mathbf{h}_t}{\partial\mathbf{W}}=\left(\dfrac{\partial\mathbf{h}_t}{\partial\mathbf{i}_t}\dfrac{\partial\mathbf{i}_t}{\partial\mathbf{e}_t}+\dfrac{\partial\mathbf{h}_t}{\partial\mathbf{f}_t}\dfrac{\partial\mathbf{f}_t}{\partial\mathbf{e}_t}+\dfrac{\partial\mathbf{h}_t}{\partial\mathbf{o}_t}\dfrac{\partial\mathbf{o}_t}{\partial\mathbf{e}_t}+\dfrac{\partial\mathbf{h}_t}{\partial\tilde{\mathbf{c}}_t}\dfrac{\partial\tilde{\mathbf{c}}_t}{\partial\mathbf{e}_t}\right)\dfrac{\partial\mathbf{e}_t}{\mathbf{W}}
\end{align*}
for $\text{LSTM}^{\text{STU}}$ and 
\begin{align*}
	\dfrac{\partial\mathbf{y}_t}{\partial\mathbf{W}}=\left(\dfrac{\partial\mathbf{y}_t}{\partial\mathbf{m}_t}\dfrac{\partial\mathbf{m}_t}{\partial\mathbf{e}_t}+\dfrac{\partial\mathbf{y}_t}{\partial\mathbf{r}_t}\dfrac{\partial\mathbf{r}_t}{\partial\mathbf{e}_t}+\dfrac{\partial\mathbf{y}_t}{\partial\tilde{\mathbf{y}}_t}\dfrac{\partial\tilde{\mathbf{y}}_t}{\partial\mathbf{e}_t}\right)\dfrac{\partial\mathbf{e}_t}{\mathbf{W}}
\end{align*}
for $\text{Highway}^{\text{STU}}$. Note that $\partial\mathbf{h}_t/\partial\mathbf{b}$, $\partial\mathbf{h}_t/\partial\mathbf{V}$, and $\partial\mathbf{y}_t/\partial\mathbf{b}$ are calculated in the same way. In addition, to use the same hyper parameters (e.g. learning rate) for both tied and untied parameters in training, the gradients of the unfolded LSTM layer parameters are further divided by the number of unfolded steps \citep{Zhang:2015ef} (here 20 unfolded steps are used). 

\section{Experimental Setup}
The proposed $\text{LSTM}^{\text{STU}}$ and $\text{Highway}^{\text{STU}}$ models were evaluated on multi-genre broadcast (MGB) data from the MGB3 speech recognition challenge task \cite{mgb3website,Bell:2015ab}. The audio is from BBC TV programmes covering a wide range of genres.
A 275 hour (275h) full training set was selected from 750 episodes 
where the training labels were from the sub-titles with a phone matched error rate $<40\%$ compared to the lightly supervised output \cite{Lanchantin:2016ab}. 
A 55 hour (55h) subset was uniformly sampled at the utterance level from the 275h set. A 63k word vocabulary \cite{Richmond:2010ab} was used with a trigram word language model (LM) estimated from both the training labels and an extra 640 million word MGB subtitle archive.
The test set, \textbf{dev17b}, contains 5.55 hours of audio data and 5,201 manually segmented utterances from 14 episodes of 13 shows. System outputs were evaluated  with 1-best Viterbi decoding as well as confusion network decoding (CN)  \cite{Mangu:2000ab,Evermann:2000ab}.

All experiments were conducted with an extended version of HTK 3.5 \cite{Young:2015ab,Zhang:2015ef}. The ANN input features were 40d log-Mel filter bank along with their 40d $\Delta$ coefficients, which were normalised at the utterance level for mean and at the show-segment level for variance \cite{Woodland:2015ab}. 
All models were trained as hybrid system acoustic models by \textit{stochastic gradient descent} based on the cross-entropy criterion with the data shuffled at the frame-level in a 800 sample minibatch \cite{Bourlard:1993ab,Dahl:2012ab,Saon:2014ab}. 
About 6k/9k decision tree clustered triphone tied-states along with appropriate training alignments were used for the 55h/275h training sets. The NewBob$^{+}$ learning rate scheduler \cite{Zhang:2015ef,Zhang:2017ab} was used for all models with the setup from our previous MGB systems \cite{Woodland:2015ab}. Weight decay factors were carefully tuned to maximise the performance of each system. 
More details about the LSTM implementation and training configuration can be found in \cite{Zhang:2018ab,Kreyssig:2018ab}.

\section{Experimental Results}

\subsection{Experiments on 55 Hour Data Set}

\subsubsection{LSTM$^{\text{STU}}$ Experiments}

The experiments started by investigating different STU settings with LSTMs. 
All 55h LSTMs had one feedforward hidden layer placed between the LSTM layers and output layer with $H=500$. Two baseline systems with one LSTM layer (1L), L$^{\text{55h}}_{0}$ and L$^{\text{55h}}_{1}$, were trained, where L$^{\text{55h}}_{0}$ was a standard LSTM and L$^{\text{55h}}_{1}$ was an LSTMP with the projection size $P=250$. 
LSTM$^{\text{STU}}$ systems with different settings were constructed: L$^{\text{55h}}_{2}$ followed those in Section~\ref{sec:LSTMSTU}; L$^{\text{55h}}_{3}$ used fixed $\bm{\eta}=\mathbf{1}$; L$^{\text{55h}}_{4}$ had untied ``peephole'' matrices; L$^{\text{55h}}_{5}$ had untied bias vectors. 
From Table~\ref{tab:55hLSTM}, L$^{\text{55h}}_{3}$ had slightly higher word error rates (WER) than L$^{\text{55h}}_{2}$, which showed generalising gating to learn $\bm{\eta}$ was useful. L$^{\text{55h}}_{4}$ outperformed L$^{\text{55h}}_{2}$ due to the use of distinct ``peepholes'', but this also increased the training difficulty and was not used. L$^{\text{55h}}_{5}$ used untied bias vectors and was found to only improve the convergence speed by producing better criterion values in the early epochs in training and similar values in the end. 
 
\begin{figure}[htb]
\vspace{-3mm}
  \centering
  \centerline{\includegraphics[width=8.5cm]{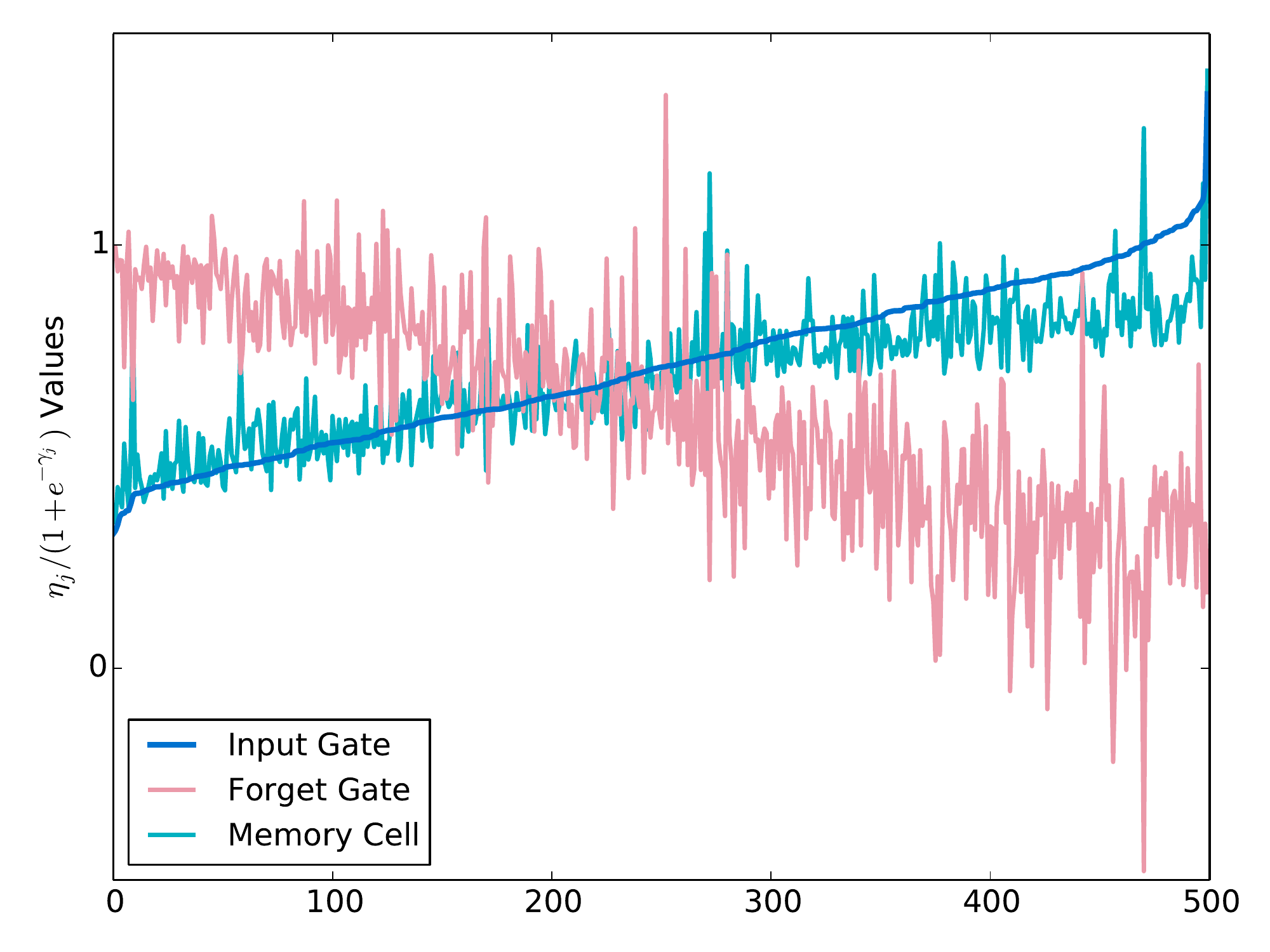}}
\vspace{-5mm}
\caption{\it 55h LSTM$^{\text{STU}}$ system L$^{\text{55h}}_{2}$ $\eta_j/(1+e^{-\gamma_j})$ values. The node indexes $j$ ($x$-axis) were re-ranked based on the input gate. }
\label{fig:lstmstu}
\vspace{-3mm}
\end{figure}
To understand the STUs in L$^{\text{55h}}_{2}$, the units used in Eqn.~\eqref{eq:5}: the input gate, forget gate and the candidate unit were shown in Fig.~\ref{fig:lstmstu}.
By ignoring $\mathbf{V}\circ\mathbf{c}_t$ and taking $\mathbf{a}_t=\mathbf{1}$, $\mathbf{i}_t$, $\mathbf{f}_t$, and $\tilde{\mathbf{c}}_t$ can be approximately evaluated by $\eta_j/(1+e^{-\gamma_j})$. 
From Fig.~\ref{fig:lstmstu}, it can be seen that the input gate and forget gate follow roughly opposite trends, which coincides with  replacing $\mathbf{i}_t$ by $\mathbf{1}-\mathbf{f}_t$ in Eqn.~\eqref{eq:11}. The candidate vector values lie in between of those from the other two units, and are more similar to $\mathbf{i}_t$ since they are multiplied to function together in Eqn.~\eqref{eq:5}. These showed that STUs can still learn reasonable gating functions with the additional untied parameters.
 
Comparing L$^{\text{55h}}_{2}$ with L$^{\text{55h}}_{0}$ and L$^{\text{55h}}_{1}$, while producing similar WERs, L$^{\text{55h}}_{2}$, L$^{\text{55h}}_{0}$, and L$^{\text{55h}}_{1}$ had 0.29 million (M), 1.16M, and 0.79M parameters in the LSTM layer. 
Hence, the use of STUs can reduce calculation and storage by a factor of four without increasing the WER. The LSTM systems with two stacked recurrent layers (2L) were also investigated, and the STU based system L$^{\text{55h}}_{8}$ still generated similar WERs to  LSTMP L$^{\text{55h}}_{7}$.
\begin{table}[h]
\centerline{
\begin{tabular}{clcc}
\toprule
ID & System & tg & cn \\
\midrule
L$^{\text{55h}}_{0}$ & 1L LSTM & 32.9 & 32.2 \\	
L$^{\text{55h}}_{1}$ & 1L LSTMP $~~~$($P=250$) & 32.9 & 32.1 \\	
L$^{\text{55h}}_{2}$ & 1L LSTM$^{\text{STU}}$  & 32.6 & 31.9 \\
L$^{\text{55h}}_{3}$ & 1L LSTM$^{\text{STU}}$ ($\bm{\eta}$ fixed to $\mathbf{1}$) & 33.3 & 32.5  \\
L$^{\text{55h}}_{4}$ & 1L LSTM$^{\text{STU}}$ (Untie $\mathbf{V}$) & 32.8 & 32.0 \\
L$^{\text{55h}}_{5}$ & 1L LSTM$^{\text{STU}}$ (Untie $\mathbf{b}$) & 33.1 & 32.2 \\	
\cmidrule{1-4}
L$^{\text{55h}}_{7}$ & 2L LSTMP $~~~$($P=250$) & 31.3 & 30.6 \\	
L$^{\text{55h}}_{8}$ & 2L LSTM$^{\text{STU}}$ & 31.4 & 30.8 \\	
\bottomrule
\end{tabular}
}
\caption{\label{tab:55hLSTM} {\it 55h LSTM system ($H=500$) \%WERs on dev17b. A trigram LM with Viterbi (tg) or CN (cn) decoding are used.}}
\vspace{-5mm}
\end{table}
\vspace{-3mm}

\subsubsection{Highway$^{\text{STU}}$ Experiments}
STUs were also used for both sigmoid and ReLU highway networks with $H=500$, and the results were listed in Table~\ref{tab:55h}. For sigmoid models, the 7 layer (7L) highway network S$^{\text{55h}}_{1}$ had a 4.2\% relative WER reduction (WERR) over the 7L deep neural network (DNN) S$^{\text{55h}}_{0}$. S$^{\text{55h}}_{1}$ and S$^{\text{55h}}_{0}$ had 4.83M and 1.61M hidden layer parameters. The Highway$^{\text{STU}}$ model, S$^{\text{55h}}_{2}$, had almost the same WERs as the standard highway network and the same number of parameters as a DNN. The use of STUs retained the WER reduction obtained from highway connections while increasing the number of hidden layer parameters by only 1.1\% rather than by 200\% with the standard highway model. The 15 layer (15L) DNN  S$^{\text{55h}}_{3}$ gave a 3.9\% WERR over the 7L DNN S$^{\text{55h}}_{0}$. 
Both standard and STU based highway systems, S$^{\text{55h}}_{4}$ and S$^{\text{55h}}_{5}$, resulted in WERRs of 3.5\% and 4.1\%  over S$^{\text{55h}}_{3}$, while using 6.51M and 0.04M extra parameters respectively.  

The same experiments were also repeated for the ReLU systems. The 7L DNN baseline R$^{\text{55h}}_{0}$ gave a 4.7\% WERR over S$^{\text{55h}}_{0}$.
2.4\% and 3.3\% WERRs were obtained by using standard and STU based highway connections. The 15L ReLU DNN, R$^{\text{55h}}_{3}$, outperformed R$^{\text{55h}}_{0}$ by a 2.4\% WERR, and its relevant highway models R$^{\text{55h}}_{4}$ and R$^{\text{55h}}_{5}$ both outperformed R$^{\text{55h}}_{3}$ by about 2\% WERR. This showed the STU idea was also applicable to ReLU. 
Note that ReLU systems obtained smaller improvements from highway connections than the sigmoid systems, which is reasonable since ReLUs suffer less from information attenuation than sigmoids.

\begin{table}[h]
\centerline{
\begin{tabular}{clcc}
\toprule
ID & System & tg & cn \\
\midrule
S$^{\text{55h}}_{0}$ & 7L sigmoid DNN & 35.8 & 34.7 \\	
S$^{\text{55h}}_{1}$ & 7L sigmoid Highway & 34.3 & 33.3 \\	
S$^{\text{55h}}_{2}$ & 7L sigmoid Highway$^{\text{STU}}$ & 34.3 & 33.2 \\	
S$^{\text{55h}}_{3}$ & 15L sigmoid DNN & 34.4 & 33.4 \\	
S$^{\text{55h}}_{4}$ & 15L sigmoid Highway & 33.2 & 32.2 \\ 
S$^{\text{55h}}_{5}$ & 15L sigmoid Highway$^{\text{STU}}$ & 33.0 & 32.0 \\ 
\cmidrule{1-4}
R$^{\text{55h}}_{0}$ & 7L ReLU DNN & 34.1 & 33.1 \\	
R$^{\text{55h}}_{1}$ & 7L ReLU Highway & 33.2 & 32.3 \\	
R$^{\text{55h}}_{2}$ & 7L ReLU Highway$^{\text{STU}}$ & 33.0 & 32.0 \\	
R$^{\text{55h}}_{3}$ & 15L ReLU DNN & 33.2 & 32.2 \\	
R$^{\text{55h}}_{4}$ & 15L ReLU Highway & 32.5 & 31.5 \\ 
R$^{\text{55h}}_{5}$ & 15L ReLU Highway$^{\text{STU}}$ & 32.6 & 31.6 \\  
\bottomrule
\end{tabular}
}
\caption{\label{tab:55h} {\it 55h highway system ($H=500$) \%WERs on dev17b. A trigram LM with Viterbi (tg) or CN (cn) decoding are used.}}
\vspace{-8mm}
\end{table}

\subsection{Experiments on 275 Hour Data Set}
In order to ensure that the 55h results and findings can scale to a significantly larger training set, some selected LSTM and highway networks were built on the full 275h set. The hidden layer size $H$ and LSTMP projection size $P$ were increased to 1000 and 500, which  quadrupled the parameters to better model the full training set. From Table~\ref{tab:275h},
the 2L LSTMP L$^{\text{275h}}_{3}$ gave a 3\% WERR over 1L LSTMP L$^{\text{275h}}_{1}$. Comparing to S$^{\text{275h}}_{2}$ and R$^{\text{275h}}_{2}$, the sigmoid and ReLU highway networks, S$^{\text{275h}}_{4}$ and R$^{\text{275h}}_{4}$, had a 4.0\% and a 3.7\% WERRs respectively by increasing the model depths from 7L to 15L.
 All of the LSTM$^{\text{STU}}$ and sigmoid/ReLU Highway$^{\text{STU}}$ systems produced similar WERs to their corresponding conventional LSTMP and highway networks while using fewer than 40\% of the parameters in the hidden layers. This validates our previous finding on a larger data set that the proposed STU can work as well as the widely used traditional gating units with far fewer parameters. The STU approach is  a highly efficient way to perform general gating and information merging that
can also be applied to other gated models, such as GRUs, recurrent highway networks, quasi-RNNs, and highway LSTMs \textit{etc}. 

\begin{table}[h]
\centerline{
\begin{tabular}{clcc}
\toprule
ID & System & tg & cn \\
\midrule
L$^{\text{275h}}_{1}$ & 1L LSTMP $~~~$($P=500$) & 26.5 & 26.0 \\	
L$^{\text{275h}}_{2}$ & 1L LSTM$^{\text{STU}}$ & 26.5 & 26.0 \\	
L$^{\text{275h}}_{3}$ & 2L LSTMP $~~~$($P=500$) & 25.7 & 25.2 \\
L$^{\text{275h}}_{4}$ & 2L LSTM$^{\text{STU}}$ & 25.9 & 25.3 \\	
\cmidrule{1-4}
S$^{\text{275h}}_{1}$ & 7L sigmoid Highway & 27.7 & 27.0 \\	
S$^{\text{275h}}_{2}$ & 7L sigmoid Highway$^{\text{STU}}$ & 27.6 & 27.0 \\	
S$^{\text{275h}}_{4}$ & 15L sigmoid Highway & 26.3 & 25.8 \\ 
S$^{\text{275h}}_{5}$ & 15L sigmoid Highway$^{\text{STU}}$ & 26.3 & 25.9 \\ 
\cmidrule{1-4}
R$^{\text{275h}}_{1}$ & 7L ReLU Highway & 27.2 & 26.4 \\	
R$^{\text{275h}}_{2}$ & 7L ReLU Highway$^{\text{STU}}$ & 27.2 & 26.3 \\ 
R$^{\text{275h}}_{4}$ & 15L ReLU Highway & 26.2 & 25.8 \\ 
R$^{\text{275h}}_{5}$ & 15L ReLU Highway$^{\text{STU}}$ & 26.1 & 25.7 \\ 
\bottomrule
\end{tabular}
}
\caption{\label{tab:275h} {\it 275h system ($H=1000$) \%WERs on dev17b. A tri-gram LM with Viterbi (tg) or CN (cn) decoding are used.}}
\vspace{-5mm}
\end{table}
\vspace{-3mm}

\section{Conclusions}

This paper proposed the use of STUs for efficient gating in LSTMs and feedforward highway networks for acoustic modelling. The weight matrices and bias vectors from all units in the same target layer are tied together to save calculations and storage space, and additional linear input/output value scaling factors are associated with the activation functions of each hidden node individually, in order to learn distinct functions for all gating and candidate units. Experiments on both 55h and 275h MGB data sets found that STU-based LSTMs and highway networks produced similar WERs to the corresponding models with traditional gating units, while being several times more efficient. It was also shown that STUs learn reasonable gating functions, 
by using only a few thousand extra untied parameters in each sublayer for gating and candidate vector generation.  


\renewcommand{\bibsection}{}


\begin{thebibliography}{10}
	
\providecommand{\newblock}{\relax}

\bibitem{Bengio:1994ab}
 Y.~Bengio, P.~Simard, \& P.~Frasconi,
 \newblock ``Learning long-term dependencies with gradient descent is difficult",
 \newblock {\em IEEE Transactions on Neural Networks},
   vol. 5, pp. 157--166, 1994.


\bibitem{Hochreiter:1997ab}
S.~Hochreiter \& J.~Schmidhuber,
 \newblock ``Long short-term memory",
 \newblock {\em Neural Computation},
   vol. 9, pp. 1735--1780, 1997.
   

\bibitem{Chung:2014ab}
J.~Chung, C.~Gulcehre, K.H.~Cho, \& Y.~Bengio,
\newblock ``Empirical evaluation of gated recurrent neural networks on sequence modeling",
\newblock  {\em arXiv.org}, 1412.3555, 2014.

\bibitem{Srivastava:2015ab}
R.K.~Srivastava, K.~Greff, \& J.~Schmidhuber,
\newblock ``Highway networks",
\newblock  {\em arXiv.org}, 1505.00387, 2015.

\bibitem{Srivastava:2015cd}
R.K.~Srivastava, K.~Greff, \& J.~Schmidhuber,
\newblock ``Training very deep networks",
\newblock  {\em Advances in NIPS 28}, Montreal, 2015.

\bibitem{Srivastava:2016ab}
J.G.~Zilly, R.K.~Srivastava, J.~Koutn\'{i}k, \& J.~Schmidhuber, 
\newblock ``Recurrent highway networks",
\newblock  {\em arXiv.org}, 1607.03474, 2016.

\bibitem{Yao:2015ab}
K.~Yao, T.~Cohn, K.~Vylomova, K. Duh, \& C.~Dyer,
\newblock ``Depth-gated {LSTM}",
\newblock  {\em arXiv.org}, 1508.03790, 2015.


\bibitem{Shi:2015ab}
X.~Shi, Z.~Chen, H.~Wang, D.-Y.~Yeung, W.-K.~Wong, \& W.-C.~Woo,
\newblock ``Convolutional {LSTM} network: {A} machine learning approach for precipitation nowcasting",
\newblock  {\em Advances in NIPS 28}, Montreal, 2015.

\bibitem{Bradbury:2017ab}
J.~Bradury, S.~Merity, C.~Xiong, \& R.~Socher,
\newblock ``Quasi-recurrent neural networks",
\newblock  {\em Proc. ICLR}, Toulon, 2017.

\bibitem{Graves:2009ab}
A.~Graves, M.~Liwicki, S.~Fernandez, R.~Bertolami, H.~Bunke, \& J.~Schmidhuber,
\newblock ``A novel connectionist system for unconstrained handwriting recognition",
 \newblock {\em IEEE Transactions on Pattern Analysis and Machine Intelligence},
   vol. 31, pp. 855--868, 2009.

\bibitem{Graves:2013ab}
A.~Graves, A.-R.~Mohamed, G.~Hinton,
\newblock ``Speech recognition with deep recurrent neural networks",
\newblock  {\em Proc. ICASSP}, Vancouver, 2013.

\bibitem{Sak:2014ab}
H.~Sak, A.~Senior, \& F.~Beaufays,
\newblock ``Long short-term memory recurrent neural network architectures for large scale acoustic modeling",
\newblock  {\em Proc. Interspeech}, Singapore, 2014.

\bibitem{Sak:2015ab}
H.~Sak, A.~Senior, K.~Rao, F.~Beaufays,
\newblock ``Fast and accurate recurrent neural network acoustic models for speech recognition",
\newblock  {\em Proc. Interspeech}, Dresden, 2015.



\bibitem{Zhang:2016ab}
Y.~Zhang, G.~Chen, D.~Yu, K.~Yao, S.~Khudanpur, \& J.~Glass,
\newblock ``Highway long short-term memory {RNN}s for distant speech recognition",
\newblock  {\em Proc. ICASSP}, Shanghai, 2016.

\bibitem{Lu:2016cd}
L.~Lu, X.~Zhang, \& S.~Renals,
\newblock ``On training the recurrent neural network encoder-decoder for large vocabulary end-to-end speech recognition",
\newblock  {\em Proc. ICASSP}, Shanghai, 2016.


\bibitem{Pundak:2017ab}
G.~Pundak \& T.N.~Sainath,
\newblock ``Highway-{LSTM} and recurrent highway networks for speech recognition",
\newblock  {\em Proc. Interspeech}, Stockholm, 2017.


\bibitem{Lu:2016ab}
L.~Lu \& S.~Renals,
\newblock ``Small-footprint deep neural networks with highway connections
for speech recognition",
\newblock  {\em Proc. Interspeech}, San Francisco, 2016.

\bibitem{Zhang:2017cd}
Y.~Zhang, W.~Chan, \& N. Jaitly,
\newblock ``Very deep convolutional networks for end-to-end speech recognition",
\newblock  {\em Proc. ICASSP}, New Orleans, 2017.

\bibitem{Tao:2017ab}
L.~Tao, Y.~Zhang, \& Y.Artzi,
\newblock ``Training {RNN}s as fast as {CNN}s",
\newblock  {\em arXiv.org}, 1709.02755, 2017.


\bibitem{Zhang:2015cd}
C.~Zhang \& P.C.~Woodland,
\newblock ``Parameterised sigmoid and {ReLU} hidden activation functions for {DNN} acoustic modelling",
\newblock  {\em Proc. Interspeech}, Dresden, 2015.

\bibitem{Zhang:2017ab}
 C.~Zhang,
\newblock {\em Joint Training Methods for Tandem and Hybrid Speech Recognition Systems using Deep Neural Networks},
\newblock  {Ph.D. thesis}, University of Cambridge, Cambridge, UK, 2017.

\bibitem{Goh:2003ab}
S.L.~Goh \& D.P.~Mandic
 \newblock ``Recurrent neural networks with trainable amplitude of activation functions",
 \newblock {\em Neural Networks},
   vol. 16, pp. 1095--1100, 2003.

\bibitem{Siniscalchi:2010ab}
S.M.~Siniscalchi, T.~Svendsen, F.~Sorbello, \& C.-H. Lee,
\newblock ``Experimental studies on continuous speech recognition using neural architectures with ``adaptive'' hidden activation functions",
\newblock  {\em Proc. ICASSP}, Dallas, 2010.

\bibitem{He:2015ab}
K.~He, X.~Zhang, S.~Ren, \& J.~Sun,
\newblock ``Delving deep into rectifiers:
{S}urpassing human-level performance on {I}mage{N}et classification",
\newblock  {\em Proc. ICCV}, Santiago, 2015.

\bibitem{Tuske:2015ab}
 Z.~T{\"u}ske, M.~Sundermeyer, R.~Schl{\"u}ter, \& H.~Ney,
 \newblock ``Integrating {G}aussian mixtures into deep neural networks: {S}oftmax layer with hidden variables",
 \newblock  {\em Proc. ICASSP}, Brisbane, 2015.

\bibitem{Siniscalchi:2013ab}
S.M.~Siniscalchi, J.~Li, \& C.-H. Lee,
\newblock ``Hermitian polynomial for speaker adaptation of connectionist speech recognition systems",
 \newblock {\em IEEE Transactions on Audio, Speech, and Language Processing},
   vol. 21, pp. 2152--2161, 2013.

\bibitem{Zhao:2015ab}
Y.~Zhao, J.~Li, J.~Xue, \& Y.~Gong,
\newblock ``Investigating online low-footprint speaker adaptation using generalized linear regression and click-through data",
\newblock  {\em Proc. ICASSP}, Brisbane, 2015.

\bibitem{Siniscalchi:2016ab}
P.~Swietojanski, J.~Li, \& S. Renals,
\newblock ``Learning hidden unit contributions for unsupervised acoustic model adaptation",
 \newblock {\em IEEE/ACM Transactions on Audio, Speech, and Language Processing},
   vol. 24, pp. 1450--1463, 2016.

\bibitem{Zhang:2016cd}
C.~Zhang \& P.C.~Woodland,
\newblock ``{DNN} speaker adaptation using parameterised sigmoid and {ReLU} hidden activation functions",
\newblock  {\em Proc. ICASSP}, Shanghai, 2016.

\bibitem{mgb3website} \url{http://www.mgb-challenge.org}

 \bibitem{Bell:2015ab}
P.~Bell, M.J.F.~Gales, T.~Hain, J.~Kilgour, P.~Lanchantin, X.~Liu, A.~McParland, S.~Renals, O.~Saz, M.~Wester, \& P.C.~Woodland,
 \newblock ``The {MGB} challenge: {E}valuating multi-genre broadcast media
transcription",
 \newblock  {\em Proc. ASRU}, Scottsdale, 2015.
 
  \bibitem{Lanchantin:2016ab}
 P.~Lanchantin, M.J.F.~Gales, P.~Karanasou, X.~Liu, Y.~Qian, L.~Wang, P.C.~Woodland, \& C.~Zhang,
  \newblock ``Selection of {M}ulti-{G}enre {B}roadcast data for the training of
 automatic speech recognition systems",
  \newblock  {\em Proc. Interspeech}, San Francisco, 2016.
  
   \bibitem{Richmond:2010ab}
  K.~Richmond, R.~Clark, \& S.~Fitt,
   \newblock ``On generating {C}ombilex pronunciations via morphological analysis",
   \newblock  {\em Proc. Interspeech}, Makuhari, 2010.
   

    \bibitem{Mangu:2000ab}
   L.~Mangu, E.~Brill, \& A.~Stolcke,
     \newblock ``Finding consensus in speech recognition: {W}ord error minimization and other applications of confusion networks",
     \newblock {\em Computer Speech \& Language},
       vol. 14, pp. 373--400, 2000.
   
    \bibitem{Evermann:2000ab}
   G.~Evermann \& P.~Woodland,
    \newblock ``Large vocabulary decoding and confidence estimation using word posterior probabilities",
    \newblock  {\em Proc. ICASSP}, Istanbul, 2000.
	
	\bibitem{Woodland:2015ab}
	P.C.~Woodland, X.~Liu, Y.~Qian, C.~Zhang, M.J.F.~Gales, P.~Karanasou, P.~Lanchantin, \& L.~Wang,
	\newblock ``Cambridge University transcription systems for the {M}ulti-{G}enre {B}roadcast challenge",
	\newblock  {\em Proc. ASRU}, Scottsdale, 2015.
	
    \bibitem{Young:2015ab}
    S.~Young, G.~Evermann, M.~Gales, T.~Hain, D.~Kershaw, X.~Liu, G.~Moore,
      J.~Odell, D.~Ollason, D.~Povey, A.~Ragni, V.~Valtchev, P.~Woodland, \& C.~Zhang,
    \newblock {\em The {HTK} Book (for {HTK} version 3.5)},
    \newblock Cambridge University Engineering Department, 2015.
	
    \bibitem{Zhang:2015ef}
    C.~Zhang \& P.C.~Woodland,
    \newblock ``A general artificial neural network extension for {HTK}",
    \newblock  {\em Proc. Interspeech}, Dresden, 2015.
	
     \bibitem{Bourlard:1993ab}
    H.A.~Bourlard \& N.~Morgan,
      \newblock ``Connectionist Speech Recognition: A Hybrid Approach",
      \newblock {Kluwer Academic Publishers}, Norwell, MA, USA 1993.
	
     \bibitem{Dahl:2012ab}
    G.E.~Dahl, D.~Yu, L.~Deng, \& A.~Acero,
      \newblock `Context-dependent pre-trained deep neural networks for large-vocabulary speech recognition",
      \newblock {\em IEEE Transactions on Audio, Speech, and Language Processing},
        vol. 20, pp. 30--42, 2012.
	
	
    \bibitem{Saon:2014ab}
    G.~Saon, H.~Soltau, A.~Emami, \& M.~Picheny,
    \newblock ``Unfolded recurrent neural networks for speech recognition",
    \newblock  {\em Proc. Interspeech}, Singapore, 2014.
	
	
    \bibitem{Zhang:2018ab}
    C.~Zhang \& P.C.~Woodland,
    \newblock ``High order recurrent neural networks for acoustic modelling",
    \newblock  {\em Proc. ICASSP}, Calgary, 2018.

 \bibitem{Kreyssig:2018ab}
F.L.~Kreyssig, C.~Zhang, \& P.C.~Woodland,
 \newblock ``Improved {TDNN}s using deep kernels and frequency dependent {G}rid-{RNN}s",
 \newblock  {\em Proc. ICASSP}, Calgary, 2018.

\end{thebibliography}
\end{document}